\documentclass[11pt,a4paper]{article}
\usepackage[hyperref]{ranlp2025}
\usepackage{times}
\usepackage{latexsym}

\usepackage{graphicx}
\usepackage{array}
\usepackage{microtype}
\usepackage{graphicx}
\usepackage{subfigure}
\usepackage{booktabs} 

\usepackage{hyperref}
\usepackage{float}
\usepackage{array}
\usepackage{amsmath}
\usepackage{amssymb}
\usepackage{mathtools}
\usepackage{amsthm}

\usepackage[capitalize,noabbrev]{cleveref}

\theoremstyle{plain}

\theoremstyle{definition}

\theoremstyle{remark}

\usepackage[textsize=tiny]{todonotes}

\usepackage{microtype}

\aclfinalcopy 

\title{Freeze and Reveal: Exposing Modality Bias in Vision-Language Models}
\author{
Vivek Hruday Kavuri, Vysishtya Karanam, Venkata Jahnavi Venkamsetty, \\
\textbf{Kriti Madumadukala, Lakshmipathi Balaji Darur, Ponnurangam Kumaraguru} \\
IIIT Hyderabad \\
\texttt{
\{kavuri.hruday, lakshmipathi.balaji\}@research.iiit.ac.in}, \\
\texttt{
\{vysishtya.karanam, venkata.venkamsetty, kriti.madumadukala\}}\\ \texttt{@students.iiit.ac.in}, \texttt{pk.guru@iiit.ac.in}
}

\date{}

\begin{document}
\maketitle
\begin{abstract}
Vision-Language Models (VLMs) achieve impressive multimodal performance but often inherit gender biases from their training data.  This bias might be coming from both the vision and text modalities. In this work, we dissect the contributions of vision and text backbones to these biases by applying targeted debiasing—Counterfactual Data Augmentation (CDA) and Task Vector methods. Inspired by data-efficient approaches in hate speech classification, we introduce a novel metric, \textit{Degree of Stereotypicality} (DoS), and a corresponding debiasing method, \textit{Data Augmentation Using DoS} (DAUDoS), to reduce bias with minimal computational cost. We curate a gender-annotated dataset and evaluate all methods on the VisoGender benchmark to quantify improvements and identify the dominant source of bias. Our results show that CDA reduces the gender gap by 6\% and DAUDoS by 3\% but using only one‑third the data. Both methods also improve the model's ability to correctly identify gender in images by 3\%, with DAUDoS achieving this improvement using only almost one-third of training data. From our experiments, we observed that CLIP's vision encoder is more biased whereas PaliGemma2's text encoder is more biased. By identifying whether the bias stems more from the vision or text encoders, our work enables more targeted and effective bias mitigation strategies in future multi-modal systems. We release our code public at \url{https://github.com/vivekhruday05/VLM_bias}
\end{abstract}
\section{Introduction}
The integration of visual and textual modalities in VLMs has led to remarkable advances in multimodal AI \cite{radford2021learningtransferablevisualmodels, steiner2024paligemma2familyversatile,blip, blip2, achiam2023gpt, team2023gemini}. VLMs have demonstrated exceptional capabilities across various tasks, including image retrieval \cite{clipvip, retrieval_sentence}, captioning \cite{blip, blip2, llava, steiner2024paligemma2familyversatile}. However these models often inherit gender biases present in their training data \cite{su-etal-2019-improving} thus making them not suitable/reliable for real world deployment. Such biases also arise from stereotypical representations in both text and images, resulting in skewed perceptions that can propagate through downstream tasks.

\begin{figure}[H]
    \centering
    \includegraphics[width=0.45\textwidth]{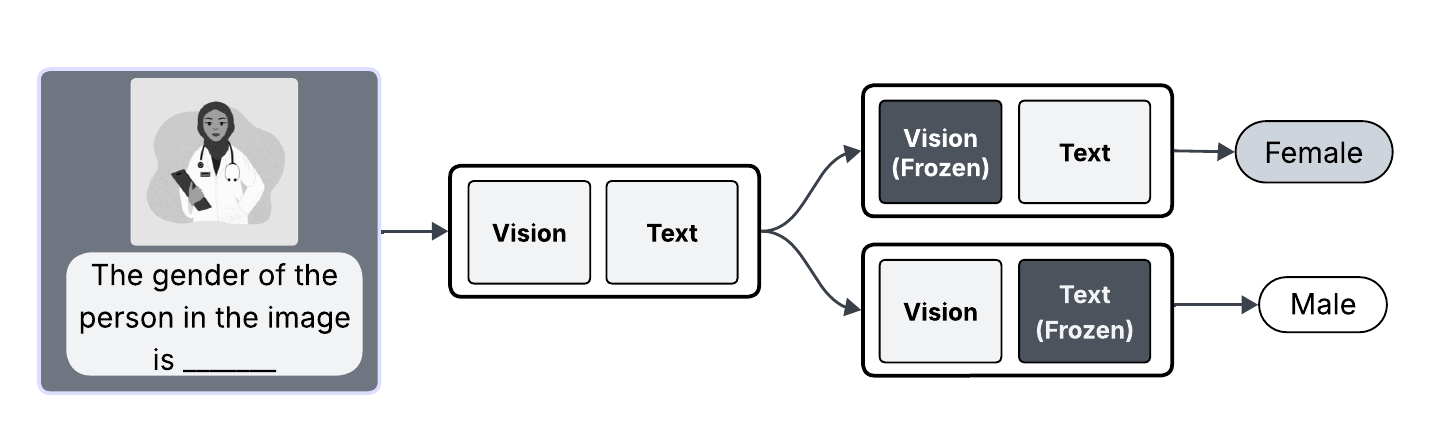}
    \caption{Different modalities posses different level of bias. We aim to show which one exhibits more bias.}
    \label{fig:Intro_pic}
\end{figure}

In this work, we address these challenges by applying targeted debiasing techniques for both modalities. Specifically for a given VLM we debias a particular modality sub-module on a curated dataset and evaluate it for gender bias using VisoGender \cite{hall2023visogenderdatasetbenchmarkinggender} to determine impact of each modality on gender bias. For this purpose we use the CelebA-Dialog dataset \cite{jiang2021talkedit} and curate the samples from the same. We annotate the data for gender based on the pronouns used in the caption and stereotypicality based on the statistical distribution of the data and insights from previous works \cite{10.3389/fpsyg.2021.733432, muthukumar2018understandingunequalgenderclassification}. To determine if a particular modality has higher influence in model's bias we evaluate it across multiple methods on our dataset. (i) We use CDA \cite{wu-dredze-2020-languages, webster2021measuringreducinggenderedcorrelations, zmigrod-etal-2019-counterfactual} a technique that mitigates bias by incorporating counterfactual data into the training process. (ii) We adapt Task Vector Unlearning \cite{dige-etal-2024-machine, ilharco2023editingmodelstaskarithmetic, zhang2023composingparameterefficientmodulesarithmetic} for debiasing. (iii) We propose a data-efficient debiasing approach, DAUDoS. We propose and do this for both CLIP-like similarity score based models and captioning type models and evaluate them across different methods. We consistently observe across multiple methods that CLIP's vision encoder is more biased compared to text encoder and in case of PaliGemma2, it's text encoder is more biased when compared to vision encoder.

In summary our key contributions are as follows:
\begin{itemize}
    \item We propose a modality‑targeted debiasing framework that applies CDA and Task‑Vector methods separately to vision and text encoders to pinpoint each modality’s bias.

    \item We curate a gender-annotated dataset for this analysis and evaluate our debiasing methods using the VisoGender benchmark.
    
    \item We propose DoS and introduce DAUDoS, lightweight debiasing methods that reduce gender bias on VisoGender with minimal overhead.

\end{itemize}
\section{Related work}

\paragraph{Bias in VLMs.}
VLMs such as CLIP and PaliGemma-2 have significantly advanced multimodal AI by integrating textual and visual modalities, enabling strong performance across diverse tasks. However, concerns have emerged regarding their tendency to inherit biases \cite{Abdollahi_2024, darur2024improvingbiasmetrics, xiao2024genderbiasemphvlbenchmarkinggenderbias, Wolfe_2023} present in training data, particularly gender bias. This bias can stem from both text and image components, as language models trained on large-scale Internet corpora frequently encode societal stereotypes, while image datasets may reinforce skewed gender representations by over representing specific demographics in certain professions, emotions, or activities. The interaction between these modalities further complicates the propagation of bias, making it crucial to determine whether textual or visual elements contribute more significantly to gender bias in VLMs. Previous works such as \cite{weng2024imagesspeaklouderwords} focus on causal mediation to trace and mitigate gender bias in GLIP, showing image features contribute most and proposing input-level blurring to reduce bias. There are also works such as \cite{srinivasan2022worstworldsbiasescompound} which deal with bias measurement to multi-modal models, revealing compounded intra and cross-modal stereotypes in VL-BERT. In contrast to these, our work targets a particular modality to find out which of the modalities contribute to a greater gender bias and whether they differ across different models and methods.

\paragraph{Bias Evaluation.}
Several studies have attempted to quantify and mitigate bias in AI models. Prior work has shown that word embeddings encode and perpetuate gender stereotypes in language representations \cite{zhao2019genderbiascontextualizedword}, that multimodal models like CLIP amplify both gender and racial biases in their image-to-text mappings \cite{Steed_2021}. There are also existing real-world benchmarks which measure societal biases in generative models, emphasizing the need for robust evaluation frameworks \cite{gehman2020realtoxicitypromptsevaluatingneuraltoxic}. Debiasing techniques focused on text prompts in multimodal models, indicating that interventions at the textual level can reduce bias to some extent but may not fully address the issue in vision-language interactions \cite{moreira2024fairpivarareducingassessingbiases}.


\paragraph{Debiasing Techniques.}
To mitigate gender bias, researchers have proposed several debiasing techniques, including CDA and Task Vector methods. CDA works by synthetically generating counterfactual training data by swapping gendered terms (e.g., replacing ``he" with ``she"), thereby balancing gender representation in textual inputs \cite{zmigrod2020counterfactualdataaugmentationmitigating} and Task Vector \cite{ilharco2023editingmodelstaskarithmetic} is an unlearning method which has it's roots originated from unlearing literature but also used in bias mitigation \cite{dige-etal-2024-machine}. While effective in NLP models, its application to VLMs remains underexplored. 

\paragraph{Data-Efficient Debiasing.}
Training on all counterfactual examples can be computationally expensive and time-consuming. To address this, prior works  \cite{nejadgholi-etal-2022-improving, garg2025ktcrimprovingimplicithate} propose approaches for improving generalization in hate speech classification while relying on fewer annotated examples. These methods leverage Concept Activation Vectors (CAVs) and introduce a novel metric, the \textit{Degree of Explicitness}, which quantifies the explicit nature of hateful content. By assigning explicitness scores to samples, they selectively fine-tune models on a curated subset of training instances, thereby enhancing efficiency without compromising performance. Inspired by these advances in NLP, we extend these ideas to the multi-modal setting and propose a novel metric termed the \textit{Degree of Stereotypicality} (DoS), which quantifies how strongly a sample exhibits stereotypical associations. Building on this, we introduce a data-efficient bias mitigation strategy called DuaDOS, which enables targeted augmentation based on stereotypicality scores. This approach reduces computational overhead while maintaining or improving model fairness and robustness in multi-modal AI systems.

\section{Dataset}
We use the CelebA-Dialog dataset \cite{jiang2021talkedit} and curate the samples from the same. This dataset contains structured annotations describing different facial attributes of celebrities and ratings of each of the attributes on a scale of 0 to 5. The captions also include gender-specific pronouns such as \textit{she}, \textit{her}, \textit{he}, \textit{him}, etc., indicating the possibility of an implicit gender labeling task. Since, we require gender for each of the data point, both for applying our methods and evaluation, we annotate the gender and describe the process in the following subsections. We also need whether a data-point is stereotypical or anti-stereotypical, so that we can use for CDA. Hence, we also annotate that attribute and describe the process in the following subsections. An example of how initial data looks like is shown in Table ~\ref{tab:initial_data}.

\begin{table}[ht]
\caption{Examples of raw dataset samples with annotations. Each image is associated with both attribute-wise and overall captions, along with a numeric rating vector indicating the prominence of each attribute (e.g., bangs, eyeglasses, beard, smile, age) in order.}
\centering
\renewcommand{\arraystretch}{1.2}
\begin{tabular}{>{\raggedleft\arraybackslash}p{0.4\columnwidth} p{0.4\columnwidth}}
\toprule
\textbf{Image} & \includegraphics[width=0.25\columnwidth]{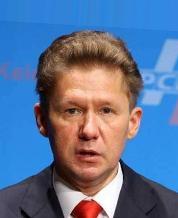} \\
\midrule
\textbf{Bangs} & He has no bangs at all. \textit{Rating: 0}\\
\textbf{Eyeglasses} & There are no eyeglasses on the face. \textit{Rating: 0}\\
\textbf{Beard} & This gentleman doesn't have any beard at all. \textit{Rating: 0}\\
\textbf{Smiling} & This gentleman looks serious with no smile on his face. \textit{Rating: 0}\\
\textbf{Age} & This person looks very old. \textit{Rating: 5}\\
\midrule
\textbf{Overall Caption} & This man in his eighties has no mustache, no fringe, and no smile. He is not wearing any eyeglasses. \\
\bottomrule
\end{tabular}
\label{tab:initial_data}
\end{table}


\subsection{Data Pre-processing and Annotation}
First, we require gender labels for every data point. To achieve this, we employ a rule-based automatic labeler. Specifically, we search for gender-related terms or pronouns such as \textit{his/her}, \textit{he/she}, \textit{gentleman/lady}, and \textit{male/female}. Based on the presence of these words, we classify the data point as male or female. If none of these words appears, the annotator assigns the label \textit{unknown}. This approach results in only 40 data points labeled as \textit{unknown}, which is negligible compared to the size of the dataset, allowing us to prune them.

Next, we annotate the data points for stereotype classification. The dataset includes a rating from 0 to 5 for each data point across attributes \{\textit{Bangs, Smiling, No Beard, Young, Eye Glasses}\}. Based on these ratings and predefined thresholds for stereotypical male and female characteristics, we label data points as either \textit{stereotypical} or \textit{anti-stereotypical}. These thresholds are determined by referring to prior publications and statistical insights from the dataset \cite{10.3389/fpsyg.2021.733432, muthukumar2018understandingunequalgenderclassification}. An example of a data point after the annotation is shown in Table ~\ref{tab:final_data}.

\begin{table}[ht]
\caption{Data sample after preprocessing. Gender and stereotype labels are added based on rule-based and attribute rating analysis, respectively. Remaining attributes such as the ratings and individual captions are discarded.}
\centering
\renewcommand{\arraystretch}{1.2}
\begin{tabular}{>{\raggedleft\arraybackslash}p{0.4\columnwidth} p{0.4\columnwidth}}
\toprule
\textbf{Image} & \includegraphics[width=0.25\columnwidth]{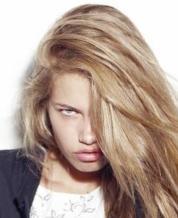} \\
\midrule
\textbf{Gender} & Female \\
\textbf{Stereotypical} & False \\
\textbf{Overall Caption} & She has no smile and no bangs.This is a young child who has no eyeglasses. \\
\bottomrule
\end{tabular}
\label{tab:final_data}
\end{table}

\begin{figure*}[h]
    \centering
    \includegraphics[width=0.75\textwidth]{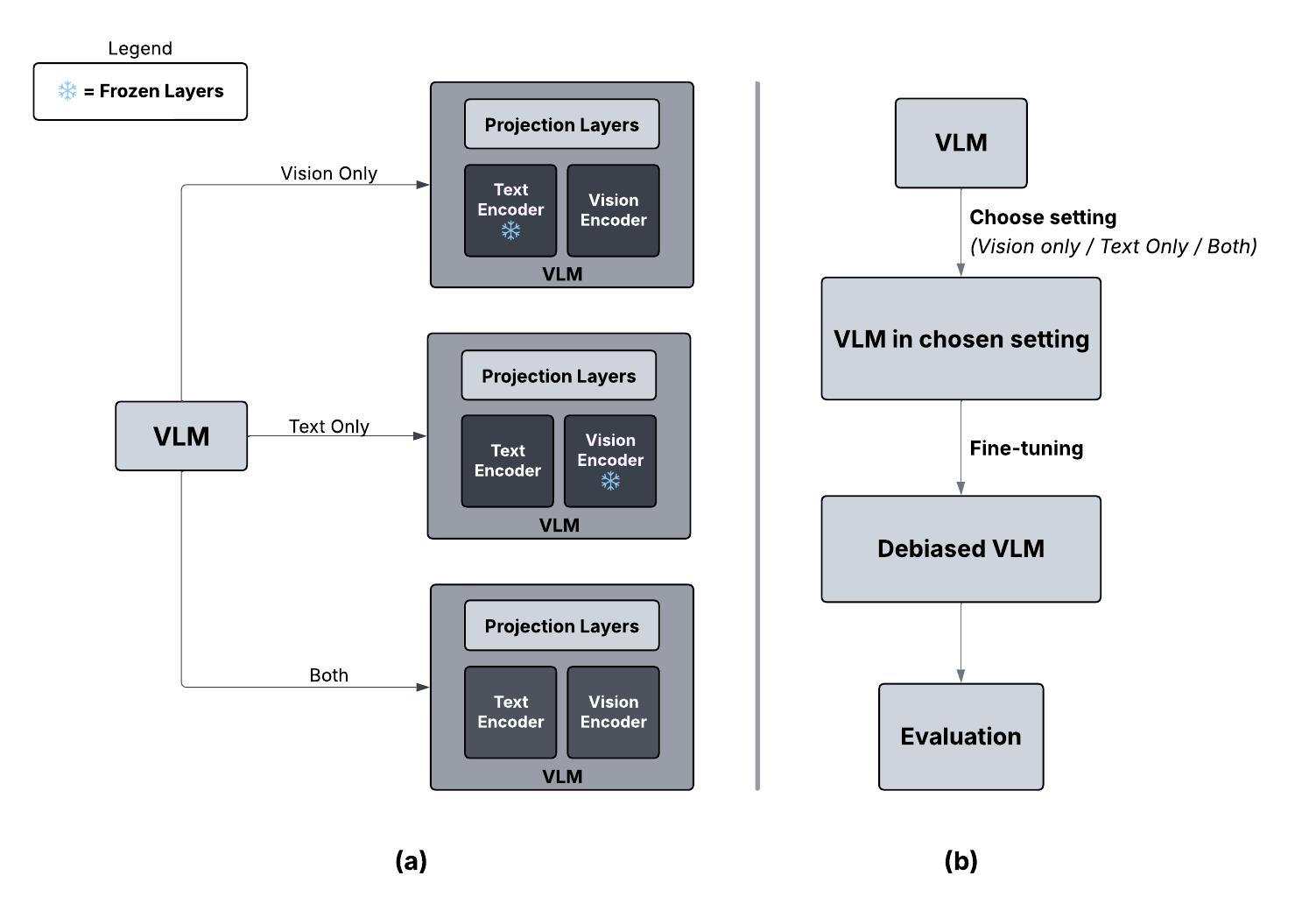}
    \caption{(a) Shows different layers that will be frozen in different settings we experiment in. (b) Shows an overall pipeline of our architecture. "Choose setting" means choosing a setting from one of the settings shown in (a).}
    \label{fig:architecture}
\end{figure*}

\section{Methodology}



Our main objective is to determine which modality—vision or text—contributes more to gender bias in our selected models. To achieve this, as shown in the Figure~\ref{fig:architecture}, we independently debias the encoder for each modality while keeping the rest of the model frozen, and then assess the overall bias using our evaluation metrics. The modality that, when debiased separately, leads to a greater reduction in bias is considered to be inherently more biased. 

This approach allows us to isolate the bias contributions of each encoder and provides insights into which modality is a more significant source of bias in the integrated VLM. To achieve this, we use pre-existing debiasing methods that debias the whole model to independently debias the encoder for each modality while keeping the rest of the model frozen. The debiasing methods we use are CDA and Weighted Task Vector.
\subsection{Counter Factual Data Augmentation}
As discussed in \cite{wu-dredze-2020-languages, webster2021measuringreducinggenderedcorrelations, zmigrod-etal-2019-counterfactual}, Counterfactual Data Augmentation (CDA) is a technique that mitigates biases by incorporating counterfactual data into the training process. In this approach, the model is fine-tuned on augmented data that challenges stereotypical associations, which helps to attenuate biased representations.

We define counterfactual data as examples that contradict prevailing stereotypes. By augmenting these anti-stereotypical examples, we hypothesize that the model will better recognize and handle non-stereotypical patterns, thus reducing inherent biases. Given that our methodology requires pre-existing debiasing mechanisms to independently address biases in the model's multimodal encoders, CDA is integrated as one of the experimental settings in our study.

\subsection{Task Vector }

As discussed in \cite{dige-etal-2024-machine, ilharco2023editingmodelstaskarithmetic, zhang2023composingparameterefficientmodulesarithmetic}, the Task Vector is derived by subtracting the weights of a base model from those of a model fine-tuned on a specific task. To enhance flexibility in debiasing strength, we introduce a \textit{weighted Task Vector method}, controlled by two hyperparameters: $\alpha$ and \texttt{blend}. Specifically, we adjust the original weights using:

\begin{equation}
W_{\text{debiased}} = W_{\text{original}} - \left((1 - \texttt{blend}) \cdot \alpha\right) \cdot \Delta W_{\text{task}}
\end{equation}

Here, $\alpha$ controls the overall intensity of debiasing, while $\texttt{blend} \in [0, 1]$ interpolates between the original and fully debiased model. A higher \texttt{blend} retains more of the original model's behavior, while a lower value emphasizes debiasing more strongly.

To identify optimal hyperparameters, we perform a random search over $\alpha \in [0.1, 1.0]$ and $\texttt{blend} \in [0.0, 1.0]$, guided by a loss that balances accuracy and fairness:

\begin{equation}
\mathcal{L} = -\text{RA}_{\text{avg}} + \lambda_{\text{gap}} \cdot \text{GenderGap}
\end{equation}

where $\text{RA}_{\text{avg}}$ is the average resolution accuracy across male and female identities, and $\text{GenderGap} = |\text{RA}_m - \text{RA}_f|$ penalizes disparity. This formulation promotes both high performance and equitable behavior by controlling for bias introduced during fine-tuning.

\subsection{Data Augmentation Using DoS (DAUDoS)}
\begin{figure}[h]
    \centering
    \includegraphics[width=0.45\textwidth]{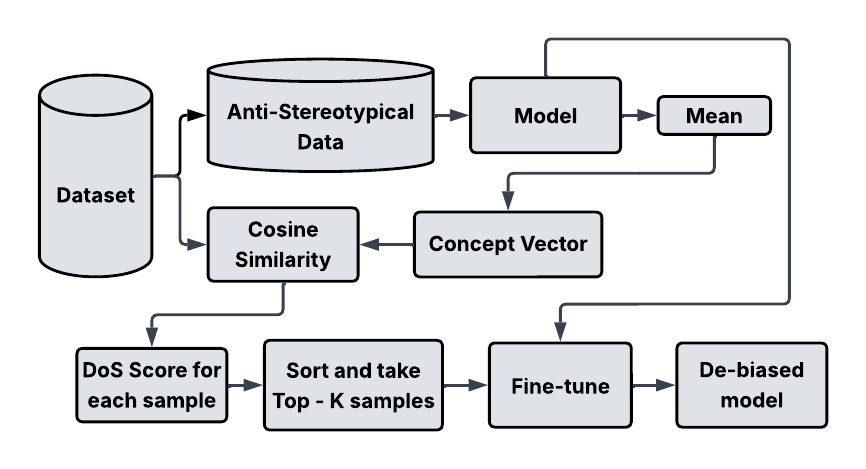}
    \caption{Depicting the method Data Augmentation Using DoS (DAUDoS). We first compute a concept vector from anti-stereotypical samples. Then, each dataset sample is scored based on its similarity to this vector, giving its Degree of Stereotypicality (DoS). The most stereotypical samples (more similarity with concept vector or score nearer to 1) are selected for fine-tuning, allowing targeted debiasing with minimal data.}
    \label{fig:DAUDoS}
\end{figure}

In this section, we introduce DAUDoS, a targeted debiasing strategy that leverages the stereotypicality of samples to perform efficient fine-tuning. The overall process is illustrated in Figure~\ref{fig:DAUDoS}.

The key idea behind DAUDoS is to assign a \textit{Degree of Stereotypicality} (DoS) score to each sample in the dataset. To do so, we begin by constructing a small set of anti-stereotypical samples. These are fed into a pre-trained model to obtain embeddings, from which we compute a \textit{Concept Activation Vector (CAV)}. Formally, if $\{\mathbf{z}_i\}_{i=1}^n$ are the model embeddings of the anti-stereotypical samples, the concept vector $\mathbf{v}_\text{CAV}$ is computed as their mean:

\begin{equation}
\mathbf{v}_\text{CAV} = \frac{1}{n} \sum_{i=1}^{n} \mathbf{z}_i.
\end{equation}

Next, for each input sample $x$, we obtain its model embedding $\mathbf{z}_x$ and compute its cosine similarity with $\mathbf{v}_\text{CAV}$:

\begin{equation}
\text{DoS}(x) = \cos(\mathbf{z}_x, \mathbf{v}_\text{CAV}).
\end{equation}

This DoS score captures how closely the sample aligns with the concept of anti-stereotypicality: higher scores indicate lower stereotypicality, and vice versa.

Once scores are assigned, we sort all training samples by their DoS values and select the top-$K$ most stereotypical samples for fine-tuning. This allows us to focus training on the subset of data that contributes most to bias, thereby making the process compute-efficient. These selected samples are used to fine-tune the model, leading to a debiased version as shown in Figure~\ref{fig:DAUDoS}.

By guiding the data augmentation process with DoS, DAUDoS minimizes training cost while retaining effectiveness in bias mitigation across modalities.

\section{Experiments}
For CDA we use the anti-stereotypical examples from the dataset we annotated and fine-tune \textit{openai/clip-vit-base-patch32}. Then for task vector, we used the stereotypical data to finetune the model and obtain task vector. In DAUDoS, we selected the samples based on the scores irrespective of what the label of the sample is (whether it is stereotypical or anti-stereotypical). We do these methods as discussed previously, in 4 different settings, namely: 

\paragraph{Text only.} In this setting, we freeze all the modules in a model except for the text encoder and projection layers related to text modality. There by only modifying the weights corresponding to the text encoder in the back propagation.
\paragraph{Vision Only.} In this setting, we freeze all the modules in a model except for the vision encoder and projection layers related to vision modality. There by only modifying the weights corresponding to the vision encoder in the back propagation.

We use Nvidia Geforce 2080 Ti for finetuning the models on the anti-stereotypical data. We describe the evaluation pipeline and the results in the upcoming sections.
\section{Results}
\begin{table*}[t]
\caption{Modality‑targeted debiasing in CLIP under OO and OP settings. High RA implies better performance, low GG implies less bias. Debiasing the vision encoder in CLIP (Vision Only) achieves the highest $RA_{\text{avg}}$ (0.97) with $GG = 0.00$, indicating vision contributes most bias.}
\label{tab:clip-oo-op-table}
\vskip 0.15in
\centering
\begin{small}
\begin{tabular}{l|cccc|cccc}
\toprule
\multicolumn{9}{c}{\textbf{CDA}} \\
\midrule
\textbf{Freeze Type} & \textbf{$RA_m$} & \textbf{$RA_f$} & \textbf{$RA_{\text{avg}}$} & \textbf{GG}
                    & \textbf{$RA_m$} & \textbf{$RA_f$} & \textbf{$RA_{\text{avg}}$} & \textbf{GG} \\
                    & \multicolumn{4}{c|}{\textbf{OO}} & \multicolumn{4}{c}{\textbf{OP}} \\
\midrule
Raw Clip        & 0.91 & 0.97 & 0.94 & 0.06 & 0.41 & 0.65 & 0.56 & 0.30 \\
Text Only   & 0.91 & \textbf{0.97} & 0.94 & 0.05 & 0.48 & 0.65 & 0.57 & 0.17 \\
Vision Only     & \textbf{0.97} & \textbf{0.97} & \textbf{0.97} & \textbf{0.00} & \textit{0.54} & 0.62 & \textit{0.58} & \textit{0.08} \\
Both     & \textbf{0.97} & \textbf{0.97} & \textbf{0.97} & \textbf{0.00} & \textbf{0.60} & \textit{0.66} & \textbf{0.63} & \textbf{0.07} \\
\midrule
\multicolumn{9}{c}{\textbf{Task Vector ($\alpha$  = 0.56, blend = 0.78)}} \\
\midrule
Text Only   & \textit{0.17} & \textbf{0.75} & \textbf{0.46} & 0.57 & 0.10 & \textit{0.02} & 0.06 & \textbf{0.08} \\
Vision Only & \textbf{0.63} & 0.23 & \textit{0.43} & \textit{0.39} & \textbf{0.56} & \textbf{0.22} & \textit{0.39} & 0.33 \\
Both     & 0.07 & \textit{0.26} & 0.17 & \textbf{0.19} & \textit{0.30} & 0.01 & 0.15 & \textit{0.29} \\
\midrule
\multicolumn{9}{c}{\textbf{DAUDoS}} \\
\midrule
Text Only    & 0.91 & \textbf{0.98} & \textit{0.95} & 0.07 & 0.38 & \textit{0.75} & 0.57 & 0.37 \\
Vision Only      & \textbf{0.94} & \textit{0.97} & \textbf{0.96} & \textbf{0.03} & \textbf{0.46} & 0.74 & \textit{0.60} & \textbf{0.29} \\
Both      & \textit{0.93} & \textbf{0.98} & \textbf{0.96} & \textit{0.05} & \textit{0.44} & \textbf{0.78} & \textbf{0.61} & 0.34 \\
\bottomrule
\end{tabular}
\end{small}
\vskip -0.1in
\end{table*}

\begin{table*}[t]
\caption{Modality‑targeted debiasing in PaliGemma2 under OO and OP settings. High RA implies better performance, low GG implies less bias. Debiasing the text encoder in PaliGemma2 (text only) yields $RA_{\text{avg}} = 0.99$ with $GG = 0.01$, showing text is the primary bias source.}
\label{tab:gemma-oo-op-table}
\vskip 0.15in
\centering
\begin{small}
\begin{tabular}{l|cccc|cccc}
\toprule
\multicolumn{9}{c}{\textbf{CDA}} \\
\midrule
\textbf{Freeze Type} & \textbf{$RA_f$} & \textbf{$RA_m$} & \textbf{$RA_{\text{avg}}$} & \textbf{GG}
                    & \textbf{$RA_f$} & \textbf{$RA_m$} & \textbf{$RA_{\text{avg}}$} & \textbf{GG} \\
                    & \multicolumn{4}{c|}{\textbf{OO}} & \multicolumn{4}{c}{\textbf{OP}} \\
\midrule
Raw Paligemma  & 0.79 & 0.46 & 0.63 & 0.33 & 0.90 & 0.45 & 0.68 & 0.45 \\
Text Only           & \textbf{0.99} & \textbf{0.98} &\textbf{ 0.99} &\textbf{ 0.01} & 0.72 & 0.78 & 0.75 & \textit{0.07} \\
Vision Only             & 0.42 & 0.39 & 0.40 & \textit{0.03} & 0.65 & 0.47 & 0.56 & 0.18 \\
Both             & 0.98 & 0.97 & 0.97 & \textbf{0.01} & \textbf{0.76} & \textbf{0.86} & \textbf{0.81} & 0.10 \\
\midrule
\multicolumn{9}{c}{\textbf{DAUDoS}} \\
\midrule
Text Only           & \textit{0.90} & \textbf{0.99} & \textit{0.94} & \textit{0.09} & \textbf{0.65} & \textit{0.87} & \textbf{0.76} & \textit{0.23} \\
Vision Only             & 0.48 & 0.67 & 0.57 & 0.19 & 0.50 & 0.80 & 0.65 & 0.30 \\
Both             & \textbf{0.93} & \textbf{0.99} & \textbf{0.96} & \textbf{0.06} & 0.52 & \textbf{0.91} & \textit{0.72} & 0.39 \\
\bottomrule
\end{tabular}
\end{small}
\vskip -0.1in
\end{table*}

To quantify gender bias in VLMs, as proposed in \cite{darur2024improvingbiasmetrics}, we employ \textbf{Resolution Accuracy (RA)} as our primary metric. RA measures the classification performance for male (\(RA_m\)) and female (\(RA_f\)) labels by evaluating how accurately the model assigns gendered labels to images. We define the \textbf{Average Resolution Accuracy} (\(RA_{avg}\)) as the mean accuracy across male and female classifications:
\begin{equation}
    RA_{avg} = \frac{RA_m + RA_f}{2}
\end{equation}
Additionally, we compute the \textbf{Gender Gap (GG)} to quantify bias intensity by measuring the difference in resolution accuracy between male and female classifications:
\begin{equation}
    GG = |RA_m - RA_f|
\end{equation}
A higher \(GG\) indicates stronger gender bias, whereas a lower \(GG\) suggests more balanced performance across genders.

Our evaluation considers model logits and their corresponding gender preferences on the Visogender benchmark \cite{hall2023visogenderdatasetbenchmarkinggender} in two settings: \textbf{Occupation-Object (OO)} and \textbf{Occupation-Participant (OP)}. 

In the \textbf{OO} setting, each instance involves a single individual paired with an occupational cue; the model is tasked with assigning the correct gender label based solely on the visual representation and the occupational context. Conversely, the \textbf{OP} setting presents a more complex scenario in which each sample includes two individuals with different roles, requiring the model to simultaneously predict the gender of multiple participants. This dual framework enables us to assess the model's ability to handle both isolated and relational gender cues, thereby providing a comprehensive view of its fairness in gender classification.

After obtaining the gender preference scores and using the true labels of the dataset, we compute \(RA_{avg}\) and \(GG\) for various debiasing configurations. In the following subsections, we report the results for the CLIP and Paligemma2 models.

\subsection{CLIP Results}

\begin{figure}[h]
    \centering
    \includegraphics[width=0.45\textwidth]{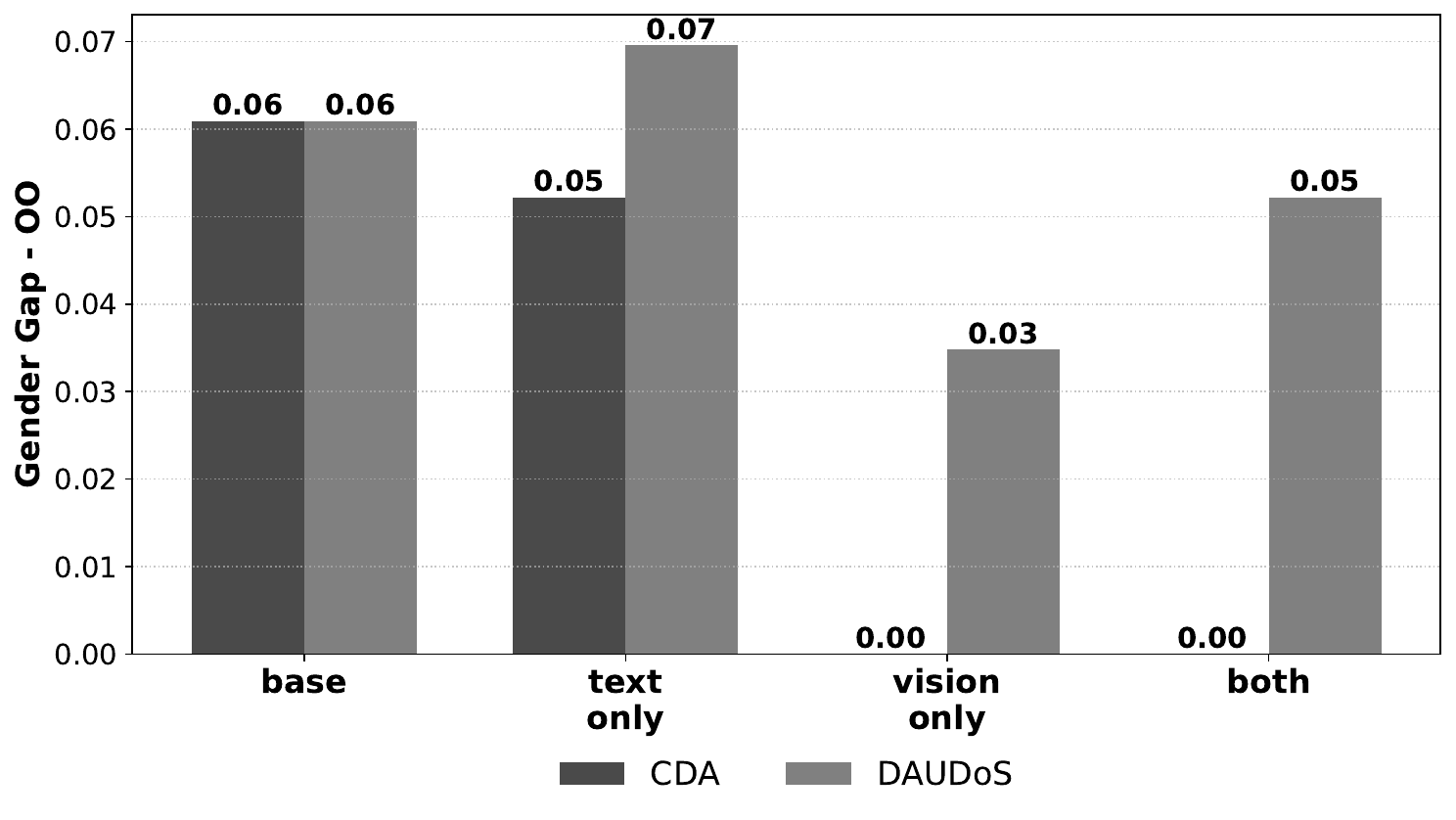}
    \caption{GG scores for \textbf{OO} setting in CLIP across debiasing configurations. Vision debiasing yields the least bias (GG = 0.0 by CDA, 0.03 by DAUDoS), similar to full model debiasing (GG = 0.0 by CDA, 0.05 by DAUDoS), indicating greater bias in the vision modality.}
    \label{fig:clip-results-oo}
\end{figure}
\begin{figure}[h]
    \centering
    \includegraphics[width=0.45\textwidth]{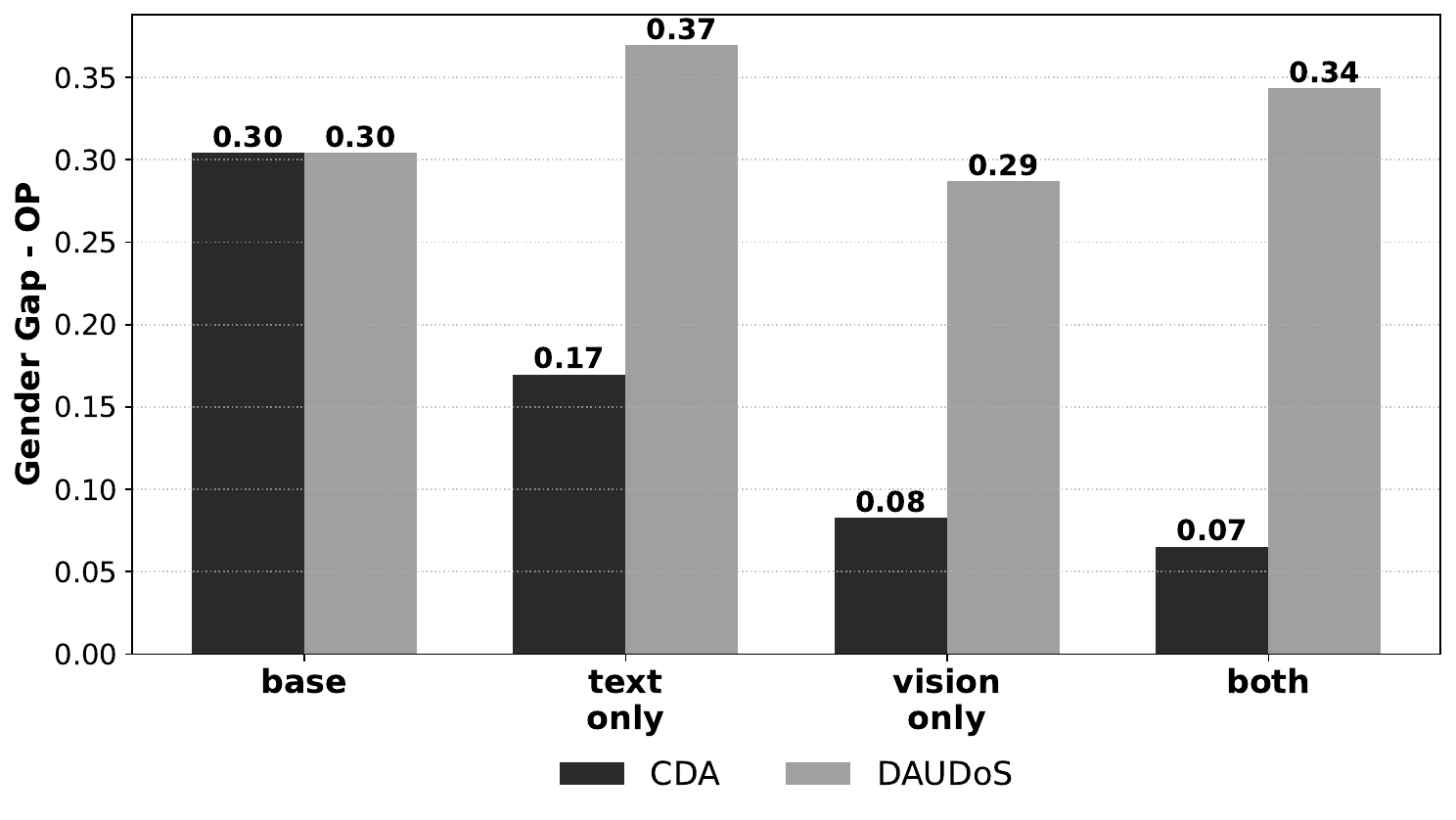}
    \caption{GG scores for \textbf{OP} setting in CLIP across debiasing configurations. Vision debiasing shows lowest bias (GG = 0.08 by CDA, 0.27 by DAUDoS), close to full model debiasing (GG = 0.07 by CDA, 0.34 by DAUDoS), again suggesting higher bias in the vision modality.}
    \label{fig:clip-results-op}
\end{figure}

Table~\ref{tab:clip-oo-op-table} summarizes the performance of CLIP under different debiasing configurations. In the OO experiments, the \emph{Raw Clip} baseline achieves an \(RA_{avg}\) of 0.94 and a moderate \(GG\) of 0.06. Debiasing the text encoder alone (text only) has almost same \(RA_{avg}\) 0.94 and decreases \(GG\) to 0.052. Notably, when the vision encoder is debiased (vision only), CLIP achieves an \(RA_{avg}\) of 0.96 with the gender gap completely eliminated (\(GG = 0.0000\)). A configuration where both encoders are left trainable (both) mirrors the outcome same as that of the case when the vision modality is debiased.

\begin{figure}[h]
    \centering
    \includegraphics[width=0.45\textwidth]{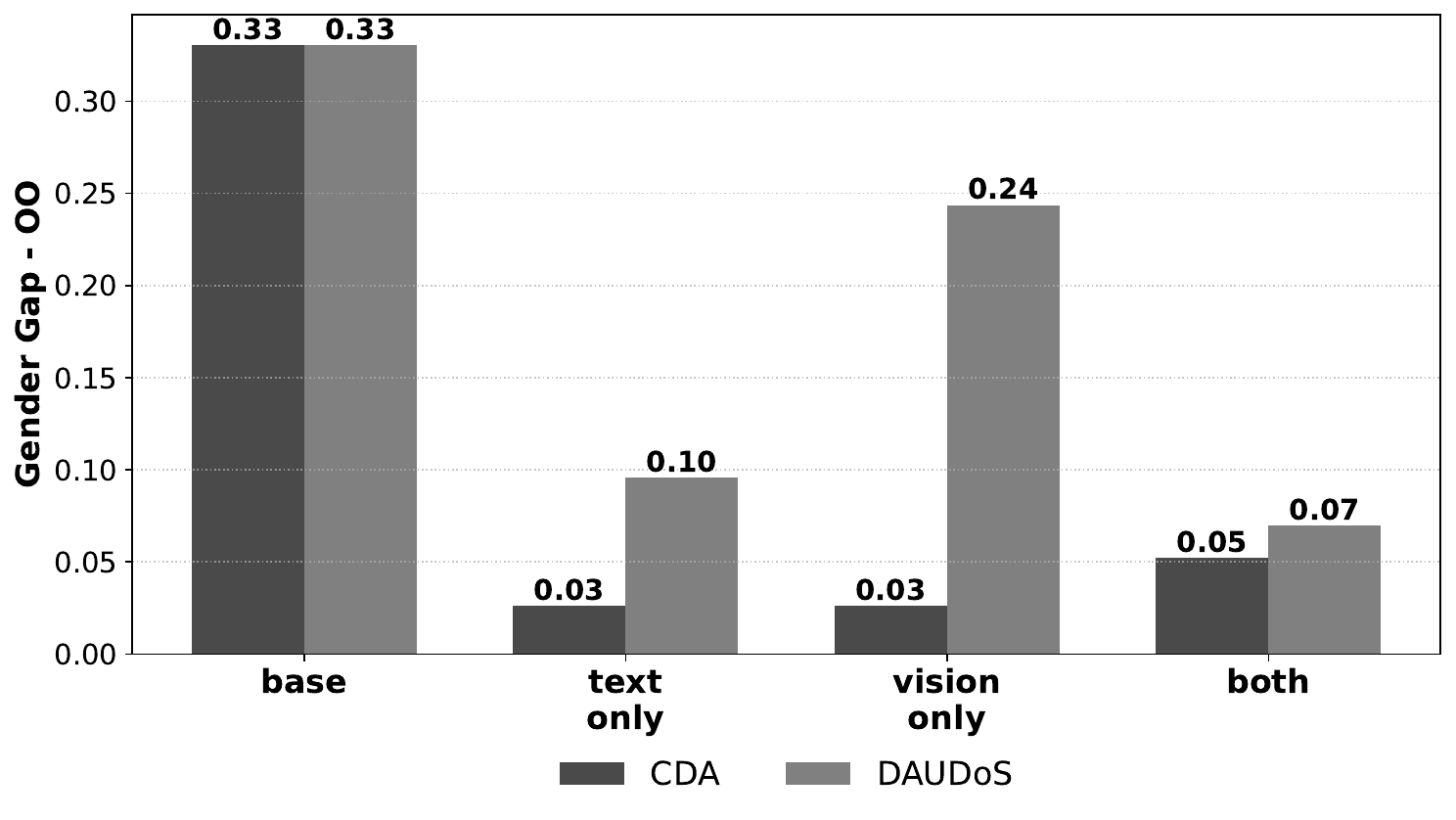}
    \caption{GG scores for \textbf{OO} setting across debiasing configurations for Paligemma2. Text debiasing yields lowest bias (GG = 0.03 by CDA, 0.10 by DAUDoS), similar to full model debiasing (GG = 0.05 by CDA, 0.07 by DAUDoS), suggesting higher bias in text modality.}
    \label{fig:gemma2-results-oo}
\end{figure}
\begin{figure}[h]
    \centering
    \includegraphics[width=0.45\textwidth]{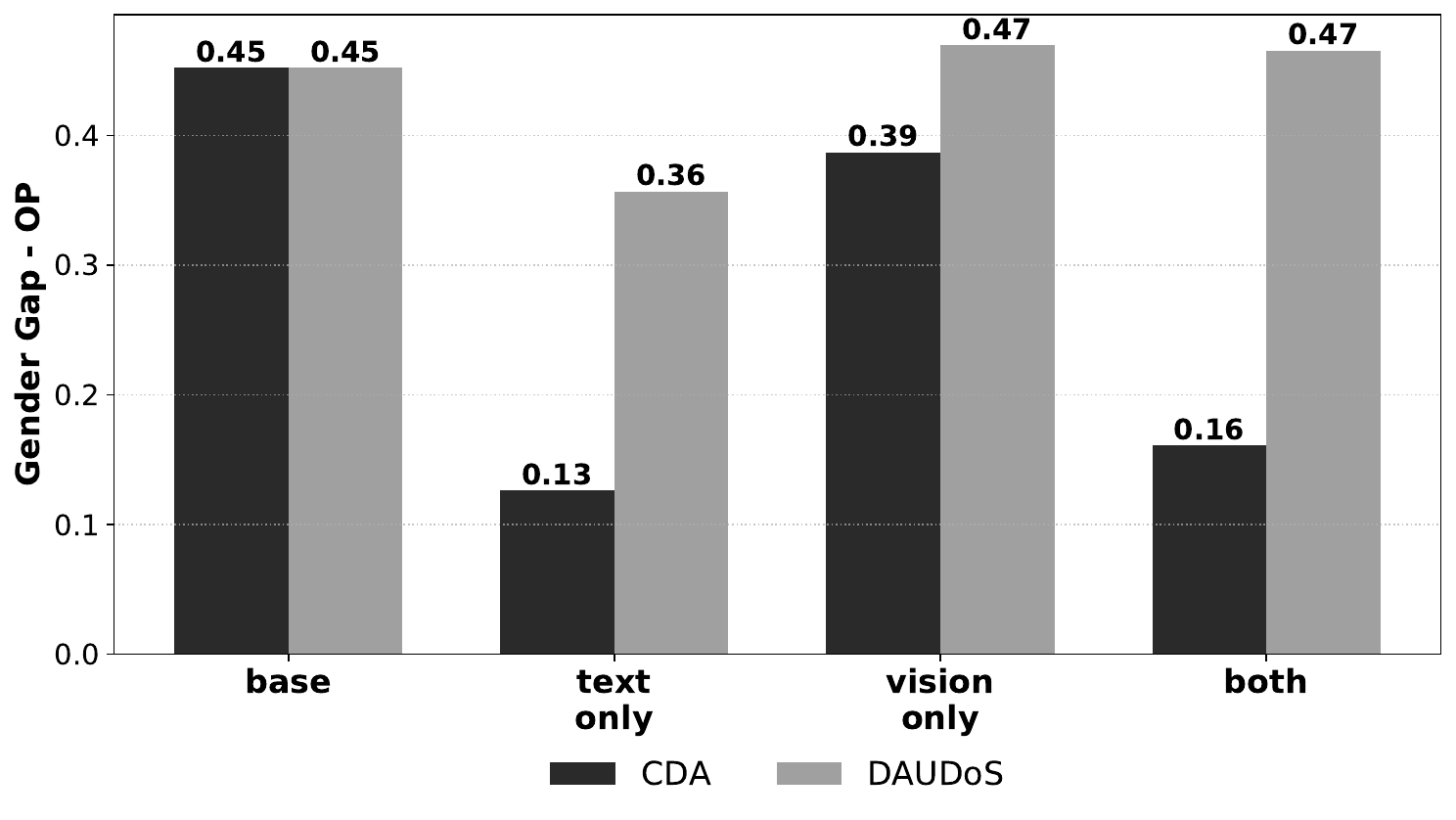}
    \caption{GG scores for \textbf{OP} setting across debiasing configurations for Paligemma2. Text debiasing gives lowest bias (GG = 0.13 by CDA, 0.36 by DAUDoS), close to full model debiasing (GG = 0.16 by CDA, 0.47 by DAUDoS), again pointing to text as the more biased modality.}
    \label{fig:gemma2-results-op}
\end{figure}

In the OP experiments (right columns of Table~\ref{tab:clip-oo-op-table}), the Raw CLIP model demonstrates a much lower accuracy compared to OO setting with \(RA_{avg}\) 0.56 and a high \(GG\) of 0.30. Debiasing the text encoder (text only) improves \(RA_{avg}\) to 0.57 and reduces \(GG\) to 0.17. Further improvement occurs when the vision encoder is debiased (vision only), yielding \(RA_{avg} = 0.58\) and \(GG = 0.08\). Finally, allowing both encoders to update (both) provides the highest \(RA_{avg}\) (0.63) with the lowest observed \(GG\) (0.06).

Figure~\ref{fig:clip-results-oo} and  Figure~\ref{fig:clip-results-op} display a plot of \(GG\) across the different debiasing configurations for CLIP, clearly illustrating that interventions aimed at debiasing the vision encoder (vision only setting) are particularly effective in lowering the gender gap. Hence, the more biased encoder in CLIP is vision encoder. We can observe this result consistently across methods.

\subsection{Paligemma2 Results}

Table~\ref{tab:gemma-oo-op-table} shows the performance of the Paligemma2 model under similar conditions. In the CDA experiments, configurations such as ``text only" and ``both" achieve very high $RA_{avg}$ (approximately 0.97--0.99) while maintaining a very low gender gap (e.g., GG=0.01 for text only). For the DAUDoS setting, while the $RA_{avg}$ remains high (around 0.94--0.96), it is important to note that these results were obtained using only one-third of the dataset. This aligns with our objective of achieving competitive performance using minimal data—demonstrating that selective sampling is both efficient and effective. Using the entire dataset would defeat the purpose of our sorting and data reduction strategy.

In the OP experiments (right columns of Table~\ref{tab:gemma-oo-op-table}), the Raw model demonstrates similar accuracy compared to OO setting with \(RA_{avg}\) 0.68 and a high \(GG\) of 0.45. Debiasing the text encoder (text only) improves \(RA_{avg}\) to 0.75 and reduces \(GG\) to 0.07. But, notably no further improvement occurs when the vision encoder is debiased (vision only), yielding \(RA_{avg} = 0.56\) and \(GG = 0.18\). Finally, allowing both encoders to update (both) provides the highest \(RA_{avg}\) (0.81) but the Gender Gap \(GG\) of (0.06) is still higher than the gender gap observed in case of text only setting.

Figure~\ref{fig:gemma2-results-oo} and Figure~\ref{fig:gemma2-results-op} provide a plot of \(GG\) for the Paligemma2 model, reinforcing the trend that debiasing the text modality (Text Only) is particularly effective in reducing gender bias. Hence the more biased modality in PaliGemma2 is the text modality. We can observe this result consistently across methods.

\section{Discussion}
Our study investigates gender bias in VLMs by independently debiasing the text and vision encoders using methods like CDA and Task Vectors. Experiments on the CelebA-Dialogue dataset and evaluations with the VisoGender benchmark reveal that targeting individual modalities is more effective than intervening at the model level. In CLIP, debiasing the vision encoder yields lower gender gaps with minimal impact on accuracy—likely due to the balanced parameter sizes across modalities. In contrast, PaliGemma2’s larger text encoder (~2.5B parameters vs. ~0.5B for vision) makes debiasing the text modality more impactful.

The findings also underscore that modality-specific debiasing leads to better bias mitigation than strategies applied post-encoder, such as projection layer adjustments, which only offer limited improvements. Our proposed DAUDoS method further supports this trend, demonstrating the generalizability of our approach across models and settings.

To conclude, we conduct experiments on the CelebA-Dialogue dataset and evaluate the outcomes using the VisoGender benchmark. Results consistently reveal that targeted debiasing of individual encoders mitigates gender bias more effectively while preserving overall model performance. By demonstrating that targeted interventions reduce gender bias while preserving performance, our work contributes practical insights for building fairer vision-language systems.

\section*{Limitations}
Despite these contributions, our study has limitations. First, the use of binary gender annotations excludes non-binary and LGBTQ+ identities, restricting the inclusiveness of our evaluation. Second, our focus is limited to gender bias and does not consider intersectional biases, such as those related to race or age.

\section*{Future Work}
In future work, we plan to broaden the scope of our analysis to address intersectional biases, such as those involving race, age, and skin tone, which may interact with gender in complex ways. This would allow for a more nuanced understanding of model fairness across diverse identities. Additionally, investigating the temporal and contextual dynamics of bias—such as how models adapt to evolving cultural norms or contextual cues can offer deeper insights into the stability and robustness of debiasing methods.

Another important direction is exploring bias mitigation strategies during the pretraining phase, rather than only through fine-tuning, to assess whether early interventions result in more systemic improvements. Finally, we plan to test our methods in real-world deployment scenarios such as image captioning, content moderation, and recommendation systems, to evaluate both fairness and utility in applied settings.

\bibliographystyle{acl_natbib}
\bibliography{anthology,ranlp2025}


\end{document}